%% file: main.tex
\documentclass[a4paper,twoside]{article}

\usepackage{epsfig}
\usepackage{subcaption}
\usepackage{calc}
\usepackage{amssymb}
\usepackage{amstext}
\usepackage{amsmath}
\usepackage{amsthm}
\usepackage{multicol}
\usepackage{pslatex}
\usepackage{apalike}
\usepackage{algorithm2e}
\usepackage[bottom]{footmisc}
\usepackage[table,xcdraw]{xcolor}

\usepackage{SCITEPRESS}     % Please add other packages that you may need BEFORE the SCITEPRESS.sty package.

\newcommand*\rot{\rotatebox{0}}

\begin{document}

\title{D-PLS: Decoupled Semantic Segmentation for 4D-Panoptic-LiDAR-Segmentation}

% \author{\authorname{A1 \sup{1}\orcidAuthor{0000-0000-0000-0000}, A2 \sup{1}\orcidAuthor{0000-0000-0000-0000} and A3\sup{2}\orcidAuthor{0000-0000-0000-0000}}

\author{\authorname{
Maik Steinhauser\sup{1},
Laurenz Reichardt\sup{1},
Nikolas Ebert\sup{1},
and Oliver Wasenm\"uller\sup{1}}
\affiliation{\sup{1}Mannheim University of Applied Sciences, Germany}
\email{\{
m.steinhauser, l.reichardt, n.ebert, o.wasenmueller\}@hs-mannheim.de
}
}

% \author{
% \authorname{Anonym Anonym\sup{1}}
% \affiliation{\sup{1} Anonym}
% \email{\{Anonym\}@xyz.edu}
% }


\abstract{
This paper introduces a novel approach to 4D Panoptic LiDAR Segmentation that decouples semantic and instance segmentation, leveraging single-scan semantic predictions as prior information for instance segmentation. Our method D-PLS first performs single-scan semantic segmentation and aggregates the results over time, using them to guide instance segmentation. The modular design of D-PLS allows for seamless integration on top of any semantic segmentation architecture, without requiring architectural changes or retraining. We evaluate our approach on the SemanticKITTI dataset, where it demonstrates significant improvements over the baseline in both classification and association tasks, as measured by the LiDAR Segmentation and Tracking Quality (LSTQ) metric. Furthermore, we show that our decoupled architecture not only enhances instance prediction but also surpasses the baseline due to advancements in single-scan semantic segmentation.
}

\onecolumn \maketitle \normalsize \setcounter{footnote}{0} \vfill

\input{Sections/1_introduction}
\input{Sections/2_related_work}
\input{Sections/3_approach}

\input{Sections/4_experiments}

\input{Sections/5_conclusions}

\section*{ACKNOWLEDGEMENTS}
This work was funded by the German Federal Ministry for Economic Affairs and Climate Action (BMWK)
under the grant SERiS
(KK5335502LB3).

\bibliographystyle{apalike}
{\small
\bibliography{references}}

\end{document}

%% file: Sections/1_introduction.tex
\begin{figure}[th!]
  \centering
  \includegraphics[width=\columnwidth]{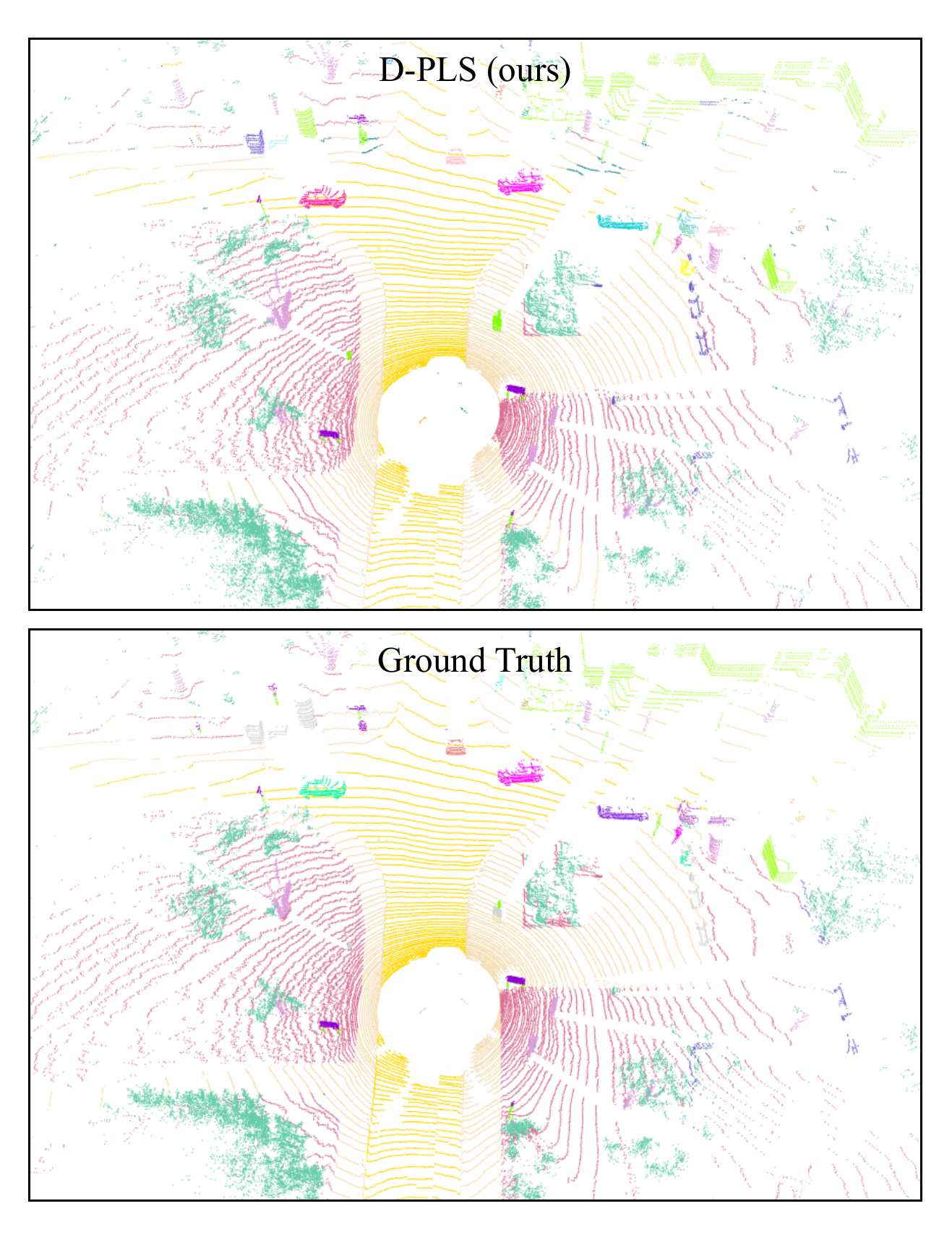}
  \caption{
  Qualitative comparison of our methods (D-PLS) results when compared to the ground-truth on the SemanticKITTI dataset.
  }
  \label{fig:teaser}
\end{figure}

\section{\uppercase{Introduction}}
\label{sec:introduction}

Light Detection and Ranging (LiDAR) sensors have become integral in providing highly accurate spatial information, enabling the 3D perception of surroundings. These capabilities make LiDAR indispensable in various applications, including autonomous navigation, mapping, robotics, and numerous industrial processes. The rise of deep learning has further enhanced the utility of LiDAR by becoming the de facto standard for processing its data, allowing for more sophisticated interpretation and application.

However, the task of 4D-Panoptic-LiDAR-Segmentation \cite{4D-PLS}, which involves segmenting LiDAR data at an instance level while maintaining temporal consistency across multiple frames, remains less explored compared to traditional LiDAR segmentation or detection methods.
As such, many new network architectures and training paradigms in the space are developed for single scan data only \cite{KPConv,SPVCNN,2DPASS,PTv3}.
However, this task is crucial for applications where understanding the evolution of objects and environments over time is essential.
In this context, we propose that semantic classes can serve as prior knowledge to improve instance segmentation by providing an initial "coarse" clustering of points.
We introduce an approach that decouples the task of 4D-Panoptic-LiDAR-Segmentation, first utilizing single scan semantic predictions as the prior information for the following step of temporal instance segmentation.
Our Decoupled Semantic Segmentation for 4D-Panoptic-LiDAR-Segmentation (D-PLS) separates temporal and instance components, offering a dedicated module on top of existing segmentation networks for analyzing dynamic environments through LiDAR data.
To assess the performance of our novel approach, we employ the SemanticKITTI \cite{SemanticKITTI} dataset. This dataset provides LiDAR point clouds and pose data for training, validation, and testing, and facilitates comparison with other state-of-the-art networks.

%% file: Sections/2_related_work.tex
\section{\uppercase{Related Work}}

The field of 4D-Panoptic-LiDAR-Segmentation has seen significant advancements, with multiple methods proposed to address the challenges of segmenting objects at an instance level with temporal consistency. The foundational work in this area, introduced by 4D-PLS \cite{4D-PLS}, established the 4D panoptic segmentation task, along with a framework for its evaluation. Using known poses from the vehicle ego motion, 4D-PLS fuses consecutive LiDAR scans into a unified spatio-temporal point cloud, which undergoes semantic segmentation. Instance clustering is performed based on predicted centers through a probabilistic framework.

Subsequent methods have built upon and extended the 4D-PLS framework. 4D-DS-Net \cite{4D-DS-Net} and 4D-StOP \cite{4D-StOP} are notable for their focus on refining the process of clustering instances by incorporating spatio-temporal proximity. 4D-DS-Net enhances the previous DS-Net \cite{DS-Net} model by introducing a dynamic shifting module that progressively adjusts individual point coordinates to move towards instance centers within the spatio-temporal context, ensuring a spatial separation and thus more accurate clustering.
Similar to our method, 4D-DS-Net decouples the semantic segmentation branch, but masks the foreground (i.e. \textit{things}) classes before passing them to the instance branch.
However, a major drawback of masking is that it eliminates spatial information and may exclude or incorrectly include misclassified points. In contrast, our method preserves all semantic information, ensuring the full spatial information in the instance branch.

4D-StOP, in contrast, adopts a different strategy by replacing the probabilistic approach of 4D-PLS with an instance-centric voting mechanism. This method involves a dual-network structure, where a voting module predicts the center of each object, and a proposal module identifies the corresponding points associated with that center. This voting-based approach provides a more precise and targeted method for determining instances over time.

Modifying the 4D-StOP method, Eq-4D-StOP \cite{EQ-4D-StOP} introduces the concept of rotation equivariance to the model. By predicting equivariant fields, this approach ensures that the model remains robust to rotational variations in the data, thereby improving the accuracy of feature learning. Instance clustering itself remains similar to 4D-StOP.

Lastly, Mask4D \cite{Mask4D} and Mask4Former \cite{Mask4Former} diverge from traditional clustering methods by utilizing a mask transformer-based paradigm. In this approach, learned queries representing potential instances are processed by attention-based decoders effectively merging segmentation and tracking into a single operation. This approach eliminates the need for non-learned clustering techniques or complex network structures, instead directly predicting spatio-temporal instance masks along with their corresponding semantic labels.

While these approaches have shown promising results, their development is decoupled from the space of semantic segmentation, with their own specific network architectures.
Adapting to the temporal context of 4D-Panoptic-LiDAR-Segmentation would require adapting the architecture or at the very least retraining.
In contrast, our work treats semantic segmentation on a single LiDAR scan as a "pre-clustering" step, which can provide valuable initial information for temporal consistency and instance level segmentation.

%% file: Sections/3_approach.tex
\begin{figure*}[t]
  \centering
  \includegraphics[width=2\columnwidth]{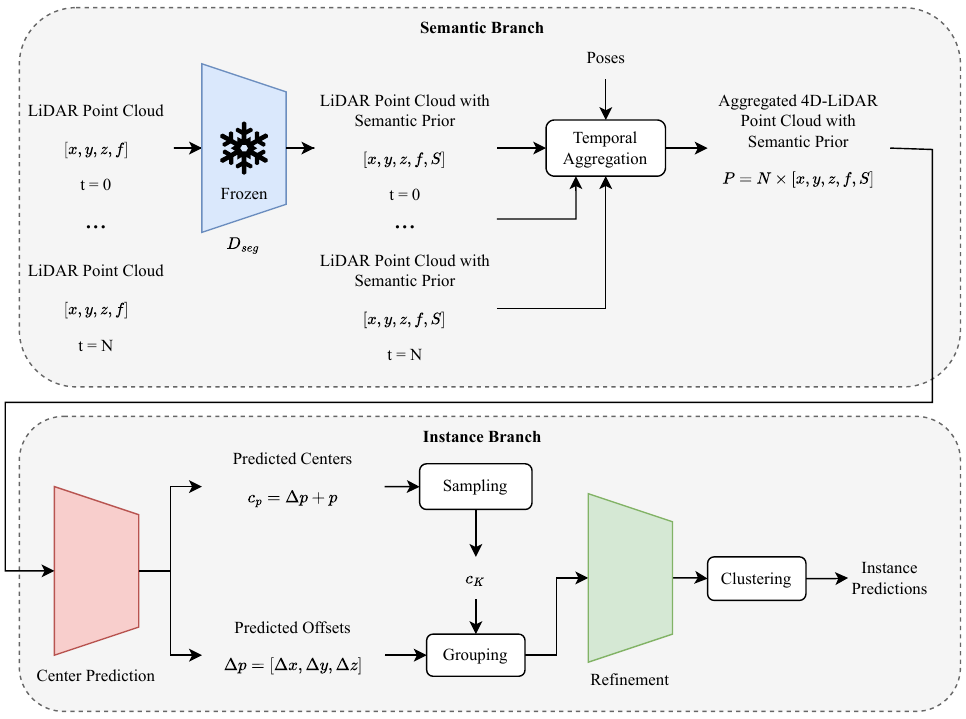}
  \caption{
  Our method D-PLS decomposes the task of 4D-Panoptic-LiDAR-Segmentation into two modular branches: the single scan semantic segmentation branch which provides prior information to the temporally aggregated instance segmentation branch.
  }
  \label{fig:method}
\end{figure*}

\section{\uppercase{Approach}}

We identify three key insights and limitations in the current state of the art for 4D-Panoptic-LiDAR-Segmentation.

Firstly, while semantic segmentation can be seen as a preliminary step to instance segmentation, where LiDAR points are initially clustered by class and then further distinguished as individual instances within each class, most methods require the network to perform both tasks simultaneously, without sharing semantic information that could enhance instance predictions.

Secondly, the few networks that do use semantic labels as the first step for instance segmentation typically only utilize them to mask out non-thing background classes such as \textit{building} or \textit{vegetation}. However, incorrect semantic predictions in these cases lead to the complete removal of geometric information for those points, depriving the instance branch of potentially valuable data.

Thirdly, developments in 4D-Panoptic-LiDAR-Segmentation often operate separately from advancements in single-scan Semantic Segmentation, with each domain focusing on distinct architectures. As a result, 4D-Panoptic-LiDAR-Segmentation does not always benefit from improvements in single-scan segmentation.

Our approach addresses these challenges by decoupling semantic segmentation from instance segmentation into two distinct steps. First we perform single-scan semantic segmentation, which acts as a coarse initial clustering of points. In a second step wen pass these temporally aggregated semantic results as prior information to aid in instance segmentation, without masking out any points. This ensures that the full geometric information is available for instance prediction. Thanks to this decoupling, our method is modular, allowing any semantic segmentation network to be used as a source of semantic prior information without requiring architectural changes or retraining (given that the desired classes remain the same).

Our method is illustrated in Figure \ref{fig:method}. The following subsections describe both steps of our method in detail.

\subsection{\uppercase{Semantic Segmentation}}
Firstly, a pre-trained single scan segmentation network $D_{seg}$ predicts semantic labels from point coordinates $x, y, z$ and their features $f$ (usually remission and/or intensity in LiDAR data) and the predicted semantic labels $S = D_{seg}(x, y, z, f)$ are appended to the original point cloud. Then, multiple 3D point clouds are temporally aggregated over $N$ time steps using ego-motion data, resulting in a 4D point cloud $P = \mathbb{R}^{N \times [x, y, z, f, S]}$.

For our experiments, we utilize the SPVCNN \cite{SPVCNN} network variant proposed in 2DPASS \cite{2DPASS}. This network combines point-based and sparse voxel-based approaches, allowing efficient 3D data processing while maintaining high resolution and modeling across various receptive field sizes. The point-based branch preserves high-resolution data, while the sparse voxel branch applies Sparse Convolution. These two branches communicate through sparse voxelization and devoxelization, with trilinear interpolation used for devoxelization. Feature fusion occurs by adding outputs from both branches, enhancing fine details with minimal computational cost. The 2DPASS variant features a lightweight decoder composed of trilinear interpolation and MLPs fusing the hierarchical features of the encoder.

This architecture is particularly effective as it maintains a pointwise branch while still leveraging hierarchical features in the voxel branch, thus avoiding the upsampling errors that occur when voxel predictions are directly upsampled as pointwise predictions. Nonetheless, any semantic segmentation architecture is compatible with our method. To preserve modularity and independence from specific architectures, we deliberately avoid using semantic embeddings, limiting the semantic prior information to one-hot predictions or confidence scores.

\subsection{\uppercase{Temporal Instance Segmentation}}

The temporally aggregated 4D point cloud, enriched with semantic prior information $P = \mathbb{R}^{N \times [x, y, z, f, S]}$, is passed to the instance segmentation branch. Here, the point cloud is processed by a hierarchical encoder-decoder architecture based on Kernel Point Convolution (KP-Conv) \cite{KPConv}, which predicts the offset $\Delta p$ between each point's $p$ coordinates and the corresponding instance's ground truth center coordinate $c_{gt}$. This allows us to coarsely assign each point to a predicted instance center$c_{p} = p + \Delta p$.
Unlike 4D-StOP, the KP-Conv network can focus solely on instance prediction.
Following 4D-StOP, we minimize the Huber loss between the predicted center $c_{p}$ and the instance center $c_{gt}$. Notably, background points such as \textit{buildings} and \textit{vegetation} are masked out using the semantic class $S$, but this masking is only applied to the loss calculation, not within the network itself.

These coarse predicted centers are further refined using the instance proposal module from 4D-StOP, which generates the final instance masks. This module first samples and groups the aggregated point cloud based on the spatial and temporal proximity of the predicted centers $c_{p}$, using Farthest Point Sampling (FPS) to select $K$ potential instance centers $c_K \subset c_p$. Points $p$ are then grouped with these centers if they fall within a predefined radius. Following this grouping, a PointNet-like module \cite{PointNet} with a multi-layer perceptron (MLP) is used to extract advanced geometric embeddings from the grouped points. These embeddings are further processed by MLPs to predict refined proposal centers, object radii, and bounding box sizes, the latter serving as an auxiliary loss to improve aggregation.
This aggregation-loss function, which incorporates errors in center position, radius, and bounding box dimensions, guides the refinement process.

Finally, DBScan \cite{DBScan} clustering is applied to the learned embeddings, aggregating points within the same cluster to form the final instance masks.
For each instance, we assign the most frequently occurring semantic label to all points within the instance.

%% file: Sections/4_experiments.tex
\input{Graphs/ablation}

\section{\uppercase{Experiments}}

Our experiments were conducted on the widely used SemanticKITTI \cite{SemanticKITTI} dataset, which is divided into training, validation, and test sets, encompassing more than $43K$ LiDAR scans captured by a Velodyne HDL-64E laser scanner in diverse urban settings. Each point within the LiDAR point clouds is annotated with one of 19 semantic categories (e.g. \textit{car}, \textit{building}, \textit{cyclist}), and foreground classes also have a unique instance ID that remains consistent over time. Additionally, the dataset provides accurate pose estimates for the ego vehicle at each time step, allowing for the spatial and temporal aggregation of point clouds, which is crucial for 4D-Panoptic-LiDAR-Segmentation.

We evaluated our method using the LiDAR Segmentation and Tracking Quality Metric (LSTQ) \cite{4D-PLS}, which assesses the performance of 4D panoptic segmentation approaches. The overall LSTQ metric is computed as the geometric mean of the classification score \( S_{\text{cls}} \) and the association score \( S_{\text{assoc}} \):

\[
LSTQ = \sqrt{S_{\text{cls}} \times S_{\text{assoc}}}
\]

The classification score \( S_{\text{cls}} \) evaluates the method’s ability to assign correct semantic labels to LiDAR points, calculated as the mean intersection over union (mIoU) across all classes. The association score \( S_{\text{assoc}} \) measures the accuracy of instance tracking throughout the entire LiDAR sequence, independent of the predicted semantic class. The geometric mean ensures that strong performance in both classification and association tasks is necessary to achieve a high overall LSTQ score.

We refer to our proposed method as D-PLS (Decoupled Panoptic LiDAR Segmentation). For the semantic prior information, we use the original checkpoint of the 2DPASS network to generate single-scan semantic predictions, which are subsequently incorporated during the temporal aggregation of the point cloud. The coarse instance segmentation stage is trained for $400K$ iterations before being frozen, followed by an additional $150K$ iterations to train the refinement stage.

\subsection{\uppercase{Ablation Study}}
\label{Sec:Ablation}

We conducted an ablation study to assess the contribution of each component in our approach.
In this study, 4D-StOP serves as the baseline, predicting both semantic and instance labels directly from a single spatio-temporally aggregated point cloud.
In contrast, our method first predicts single-scan semantic labels using a semantic network before aggregating point clouds across multiple time steps.
We evaluate whether this semantic prior information is best added as one-hot semantics or as semantic confidences.

We also experiment with having the instance stage of our method being supervised by an auxiliary semantic segmentation loss, in order to further refine the semantic prior going into the network.
The reasoning behind this experiment is to determine whether predicting both semantic and instance information is detrimental.

Table \ref{Table:Ablation} presents the results, where $N=2$ scans were aggregated spatio-temporally during training and evaluation. Note that the baseline results differ from the original 4D-StOP paper, as we reduced the maximum points per batch to accommodate our hardware limitations. As a result, both the baseline and our method were retrained under identical conditions to ensure a fair comparison. Our findings indicate that introducing one-hot semantics as prior information for instance segmentation results in a clear improvement in LSTQ compared to the baseline. In particular, the classification score \( S_{\text{cls}} \) saw significant enhancement, demonstrating that network architectures and training strategies developed for single-scan data can be highly beneficial in the context of panoptic segmentation. This suggests that future advancements in single-scan segmentation could further improve panoptic segmentation.

Moreover, the use of semantic predictions as prior information has positively impacted the association score \( S_{\text{assoc}} \), leading to improved instance predictions.
Our hypothesis that decoupling semantic and instance predictions benefits panoptic segmentation is further supported by the observation that retaining the auxiliary semantic loss negatively affects the instance segmentation stage.
We use the configuration leading to the best result for comparison with other state-of-the-art methods in the next section.

\section{\uppercase{Comparison with State of the Art}}

For our final evaluation, we compared our approach with the current state-of-the-art methods.
Based on the results from the ablation study in Section \ref{Sec:Ablation}, we use one-hot semantics as the semantic prior and we retrained the model with $N=4$ spatio-temporally aggregated scans. This setting allows for equal comparability with the state of the art. The results, shown in Table \ref{Table:SOTA}, demonstrate the effectiveness of our method, which outperforms the state of the art, except for Mask4D.
However, unlike our approach, Mask4D is not modular, meaning that integrating new developments from single-scan applications requires adaptations and retraining.

\input{Graphs/sota}

Our method shows a substantial improvement over the baseline 4D-StOP, confirming the viability of decoupling semantic segmentation from instance mask prediction. This concept could also be applied to methods like Mask4D, with further improvements possible by incorporating recent advancements in semantic segmentation networks such as Point Transformer V3 \cite{PTv3}. We leave this for future research.
Notably, when compared to 4D-DS-Net which uses semantic results to mask out points prior to instance refinement, our method performs better, indicating that our approach of not using semantic masking in the instance branch leads to better results.

%% file: Graphs/ablation.tex
\begin{table*}[t]
\centering
\caption{
Ablation study of the components of our method on the SemanticKITTI validation set using $N=2$ scans.
}
\resizebox{2\columnwidth}{!}{

\begin{tabular}{cccccccc}
\hline
\rowcolor[HTML]{EFEFEF} 
\rot{\textbf{One-Hot Semantics}} & \rot{\textbf{Semantic Confidences}} & \rot{\textbf{Auxiliary Semantic Loss}} & \textbf{$LSTQ \%$} & \textbf{$S_{assoc} \%$} & \textbf{$S_{cls} \%$} & \textbf{$IoU_{Th} \%$} & \textbf{$IoU_{St} \%$} \\ \hline
\multicolumn{3}{c|}{none (Baseline) \cite{4D-StOP}}                                                          & 58.01            & 65.50              & 51.38            & 45.15             & 60.86             \\
\checkmark                        &                               & \multicolumn{1}{c|}{}            & \textbf{69.50}   & 73.00              & 66.17            & 69.93             & 69.45             \\
\checkmark                        &                               & \multicolumn{1}{c|}{\checkmark}         & 68.71            & 71.42              & 66.10            & 69.75             & 69.45             \\
                           & \checkmark                           & \multicolumn{1}{c|}{}            & 69.26            & 72.53              & 66.13            & 69.84             & 69.45             \\
                           & \checkmark                           & \multicolumn{1}{c|}{\checkmark}         & 69.34            & 72.63              & 66.20            & 70.00             & 69.45             \\ \hline
\end{tabular}

}
\label{Table:Ablation}
\end{table*}

%% file: Graphs/sota.tex
\begin{table}[h!]
\centering
\caption{
Comparison of our method using $N=4$ scans compared to others on the SemanticKITTI validation set.
}
\resizebox{\columnwidth}{!}{

\begin{tabular}{cccccc}
\hline
\rowcolor[HTML]{EFEFEF} 
\textbf{Method}                  & \textbf{$LSTQ \%$} & \textbf{$S_{assoc} \%$} & \textbf{$S_{cls} \%$} & \textbf{$IoU_{Th} \%$} & \textbf{$IoU_{St} \%$} \\ \hline
\multicolumn{1}{c|}{4DPLS \cite{4D-PLS}}       & 59.86            & 58.79              & 60.95            & 64.96             & 63.06             \\

\multicolumn{1}{c|}{CA-Net \cite{CA-Net}}       & -            & 72.90              & -            & -             & -             \\

\multicolumn{1}{c|}{4DStOP \cite{4D-StOP}}      & 66.40            & 71.80              & 61.40            & 64.90             & 64.10             \\
\multicolumn{1}{c|}{Eq-4D-PLS \cite{EQ-4D-StOP}}   & 65.00            & 67.70              & 62.30            & 64.60             & 66.40             \\
\multicolumn{1}{c|}{4D-DS-Net \cite{4D-DS-Net}}   & 68.00            & 71.30              & 64.80            & 64.50             & 65.30             \\
\multicolumn{1}{c|}{Mask4D \cite{Mask4D}}      & 71.40            & 75.40              & 67.50            & 65.80             & 69.90             \\
\multicolumn{1}{c|}{Eq-4D-StOP \cite{EQ-4D-StOP}}  & 70.10            & 77.60              & 63.40            & 67.10             & 66.40             \\
\multicolumn{1}{c|}{Mask4Former \cite{Mask4Former}} & 70.50            & 74.30              & 66.90            & 67.10             & 66.60             \\
\multicolumn{1}{c|}{ours}        & 70.49            & 74.87              & 66.36            & 70.40             & 69.45             \\ \hline
\end{tabular}

}
\label{Table:SOTA}
\end{table}

%% file: Sections/5_conclusions.tex
\section{\uppercase{Conclusion}}
\label{sec:conclusion}

This paper introduces a novel two-stage approach to 4D Panoptic Lidar Segmentation by decoupling single-scan semantic and multi-scan instance segmentation tasks.
Our D-PLS framework leverages semantic predictions as a coarse form of clustering and prior information to aid in instance segmentation, significantly outperforming the baseline, while offering a modular and flexible solution.
D-PLS capitalizes on pretrained networks from the well-explored domain of single-scan semantic segmentation to supply these prior predictions, making it adaptable to a wide range of architectures. Future work will explore extending our approach to diverse semantic and panoptic networks. We validated our results using the SemanticKITTI dataset, and aim to further advance this research by incorporating label-efficient LiDAR segmentation techniques \cite{ImageTo360}, instance specific augmentation \cite{Text3DAug} and motion information \cite{rishav2020deeplidarflow}.

%% file: main.bbl
\begin{thebibliography}{}

\bibitem[Ayg\"un et~al., 2021]{4D-PLS}
Ayg\"un, M., Osep, A., Weber, M., Maximov, M., Stachniss, C., Behley, J., and Leal-Taixe, L. (2021).
\newblock {4D Panoptic Segmentation}.
\newblock In {\em Conference on Computer Vision and Pattern Recognition (CVPR)}.

\bibitem[Behley et~al., 2019]{SemanticKITTI}
Behley, J., Garbade, M., Milioto, A., Quenzel, J., Behnke, S., Stachniss, C., and Gall, J. (2019).
\newblock {SemanticKITTI: A Dataset for Semantic Scene Understanding of LiDAR Sequences}.
\newblock In {\em International Conference on Computer Vision (ICCV)}.

\bibitem[Ester et~al., 1996]{DBScan}
Ester, M., Kriegel, H.-P., Sander, J., Xu, X., et~al. (1996).
\newblock A density-based algorithm for discovering clusters in large spatial databases with noise.
\newblock In {\em KDD}.

\bibitem[Hong et~al., 2021]{DS-Net}
Hong, F., Zhou, H., Zhu, X., Li, H., and Liu, Z. (2021).
\newblock Lidar-based panoptic segmentation via dynamic shifting network.
\newblock In {\em Conference on Computer Vision and Pattern Recognition (CVPR)}.

\bibitem[Hong et~al., 2022]{4D-DS-Net}
Hong, F., Zhou, H., Zhu, X., Li, H., and Liu, Z. (2022).
\newblock Lidas-based 4d panoptic segmentation via dynamic shifting network.
\newblock {\em arXiv preprint arXiv:2203.07186}.

\bibitem[Kreuzberg et~al., 2022]{4D-StOP}
Kreuzberg, L., Zulfikar, I.~E., Mahadevan, S., Engelmann, F., and Leibe, B. (2022).
\newblock 4d-stop: Panoptic segmentation of 4d lidar using spatio-temporal object proposal generation and aggregation.
\newblock In {\em European Conference on Computer Vision (ECCV)}.

\bibitem[Marcuzzi et~al., 2023]{Mask4D}
Marcuzzi, R., Nunes, L., Wiesmann, L., Marks, E., Behley, J., and Stachniss, C. (2023).
\newblock Mask4d: End-to-end mask-based 4d panoptic segmentation for lidar sequences.
\newblock {\em Robotics and Automation Letters (RA-L)}.

\bibitem[Marcuzzi et~al., 2022]{CA-Net}
Marcuzzi, R., Nunes, L., Wiesmann, L., Vizzo, I., Behley, J., and Stachniss, C. (2022).
\newblock Contrastive instance association for 4d panoptic segmentation using sequences of 3d lidar scans.
\newblock {\em Robotics and Automation Letters (RA-L)}.

\bibitem[Qi et~al., 2017]{PointNet}
Qi, C.~R., Su, H., Mo, K., and Guibas, L.~J. (2017).
\newblock Pointnet: Deep learning on point sets for 3d classification and segmentation.
\newblock In {\em Conference on Computer Vision and Pattern Recognition}.

\bibitem[Reichardt et~al., 2023]{ImageTo360}
Reichardt, L., Ebert, N., and Wasenm{\"u}ller, O. (2023).
\newblock 360${}^\circ$ from a single camera: A few-shot approach for lidar segmentation.
\newblock In {\em International Conference on Computer Vision Workshop (ICCVW)}.

\bibitem[Reichardt et~al., 2024]{Text3DAug}
Reichardt, L., Uhr, L., and Wasenm{\"u}ller, O. (2024).
\newblock Text3daug – prompted instance augmentation for lidar perception.
\newblock In {\em International Conference on Intelligent Robots and Systems (IROS)}.

\bibitem[Rishav et~al., 2020]{rishav2020deeplidarflow}
Rishav, R., Battrawy, R., Schuster, R., Wasenm{\"u}ller, O., and Stricker, D. (2020).
\newblock Deeplidarflow: A deep learning architecture for scene flow estimation using monocular camera and sparse lidar.
\newblock In {\em IEEE/RSJ International Conference on Intelligent Robots and Systems (IROS)}.

\bibitem[Tang et~al., 2020]{SPVCNN}
Tang, H., Liu, Z., Zhao, S., Lin, Y., Lin, J., Wang, H., and Han, S. (2020).
\newblock Searching efficient 3d architectures with sparse point-voxel convolution.
\newblock In {\em European Conference on Computer Vision (ECCV)}.

\bibitem[Thomas et~al., 2019]{KPConv}
Thomas, H., Qi, C.~R., Deschaud, J.-E., Marcotegui, B., Goulette, F., and Guibas, L.~J. (2019).
\newblock Kpconv: Flexible and deformable convolution for point clouds.
\newblock In {\em International Conference on Computer Vision (ICCV)}.

\bibitem[Wu et~al., 2024]{PTv3}
Wu, X., Jiang, L., Wang, P.-S., Liu, Z., Liu, X., Qiao, Y., Ouyang, W., He, T., and Zhao, H. (2024).
\newblock Point transformer v3: Simpler faster stronger.
\newblock In {\em Conference on Computer Vision and Pattern Recognition (CVPR)}.

\bibitem[Yan et~al., 2022]{2DPASS}
Yan, X., Gao, J., Zheng, C., Zheng, C., Zhang, R., Cui, S., and Li, Z. (2022).
\newblock 2dpass: 2d priors assisted semantic segmentation on lidar point clouds.
\newblock In {\em European Conference on Computer Vision (ECCV)}.

\bibitem[Yilmaz et~al., 2024]{Mask4Former}
Yilmaz, K., Schult, J., Nekrasov, A., and Leibe, B. (2024).
\newblock Mask4former: Mask transformer for 4d panoptic segmentation.
\newblock In {\em International Conference on Robotics and Automation (ICRA)}.

\bibitem[Zhu et~al., 2023]{EQ-4D-StOP}
Zhu, M., Han, S., Cai, H., Borse, S., Ghaffari, M., and Porikli, F. (2023).
\newblock 4d panoptic segmentation as invariant and equivariant field prediction.
\newblock In {\em International Conference on Computer Vision (ICCV)}.

\end{thebibliography}
